\documentclass[letterpaper, 10 pt, journal, twoside]{IEEEtran}
\usepackage{hhline}
\usepackage{graphicx}
\usepackage{adjustbox}
\usepackage{subcaption}
\usepackage{multirow}
\usepackage{soul,color}
\usepackage{amsfonts}
\usepackage{float}
\usepackage{amsmath}
\usepackage{algorithm}
\usepackage{algorithmic}
\usepackage{boldline} 
\usepackage{xcolor,pifont}
\usepackage{tikz}
\usepackage{amssymb}
\usepackage{pifont}
\newcommand*\colourx[1]{%
  \expandafter\newcommand\csname #1x\endcsname{\textcolor{#1}{\ding{55}}}%
}
\colourx{blue}
\colourx{green}
\colourx{red}

\newcommand*\colourcheck[1]{%
  \expandafter\newcommand\csname #1check\endcsname{\textcolor{#1}{\ding{51}}}%
}
\colourcheck{blue}
\colourcheck{green}
\colourcheck{red}

\usepackage{kotex}
\usepackage{kotex-logo}
\usepackage{kotex}
\usepackage{kotex}
\usepackage{kotex-logo}
\usepackage{kotex}
\IEEEoverridecommandlockouts                              

\title{Training Robots without Robots: Deep Imitation Learning for Master-to-Robot Policy Transfer
}
\author{Heecheol Kim$^{1}$$^{,}$$^{3}$, Yoshiyuki Ohmura$^{1}$, Akihiko Nagakubo$^{2}$, and Yasuo Kuniyoshi$^{1}$%
\thanks{Manuscript received: September, 29, 2022; Revised January, 5, 2023; Accepted March, 12, 2023.}
\thanks{This paper was recommended for publication by Editor Dana Kulic upon evaluation of the Associate Editor and Reviewers' comments. 
This paper is partly based on results obtained under a Grant-in-Aid for Scientific Research (A) JP22H00528 and supported in part by the Department of Social Cooperation Program ``Intelligent Mobility Society Design,'' funded by Toyota Central R\&D Labs., Inc.,  of the Next Generation AI Research Center, The University of Tokyo.} 
\thanks{$^{1}$Heecheol Kim, Yoshiyuki Ohmura, and Yasuo Kuniyoshi are with Laboratory for Intelligent Systems and Informatics, Graduate School of Information Science and Technology, The University of Tokyo, 7-3-1 Hongo, Bunkyo-ku, Tokyo, Japan
        {\tt\footnotesize \{h-kim, ohmura, kuniyosh\}@isi.imi.i.u-tokyo.ac.jp}}%
\thanks{$^{2}$ Akihiko Nagakubo is with Artificial Intelligence Research Center, National Institute of Advanced Industrial Science and Technology, 1-1-1 Umezono, Tsukuba, Ibaraki, Japan {\tt\footnotesize nagakubo.a@aist.go.jp}}
\thanks{$^{3}$ Corresponding author}
\thanks{Digital Object Identifier (DOI): see top of this page.}
}

\begin{document}
\maketitle


\begin{abstract}
Deep imitation learning is promising for robot manipulation because it only requires demonstration samples. In this study, deep imitation learning is applied to tasks that require force feedback. However, existing demonstration methods have deficiencies; bilateral teleoperation requires a complex control scheme and is expensive, and kinesthetic teaching suffers from visual distractions from human intervention.
This research proposes a new master-to-robot (M2R) policy transfer system that does not require robots for teaching force feedback-based manipulation tasks.
The human directly demonstrates a task using a controller. This controller resembles the kinematic parameters of the robot arm and uses the same end-effector with force/torque (F/T) sensors to measure the force feedback. 
Using this controller, the operator can feel force feedback without a bilateral system. The proposed method can overcome domain gaps between the master and robot using gaze-based imitation learning and a simple calibration method. Furthermore, a Transformer is applied to infer policy from F/T sensory input. The proposed system was evaluated on a bottle-cap-opening task that requires force feedback.

\end{abstract}


\section{Introduction}

\begin{figure*}
\centering 
\vspace{0.1in}
\includegraphics[width=.65\linewidth]{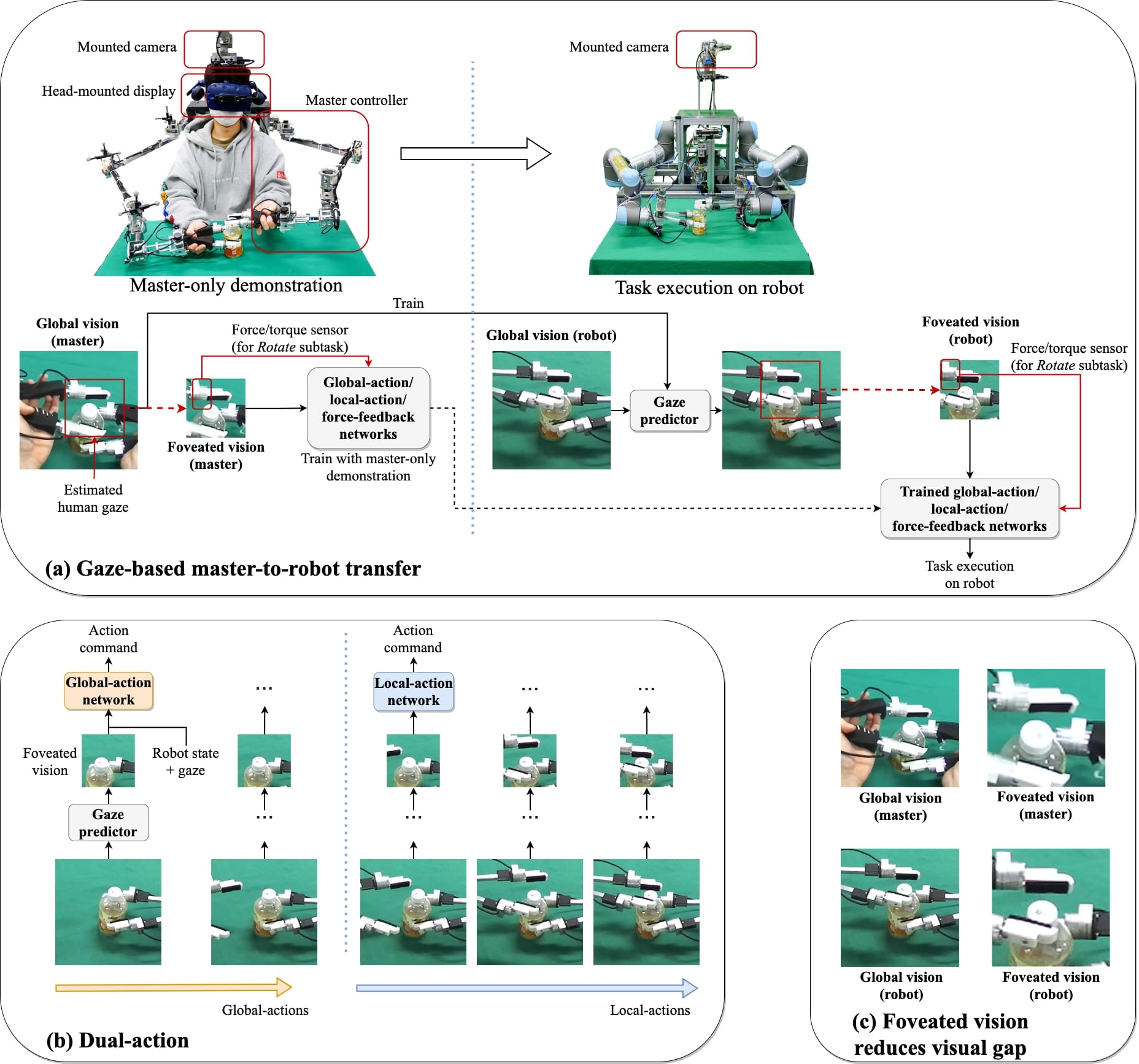}
\caption{(a) 
Master-to-robot (M2R) policy transfer system can teach robots tasks that require force feedback from master-only-generated demonstrations. 
This method uses gaze-based visual attention to minimize the visual gap between the master and robot. 
(b) The dual-action \cite{kim2021gaze}, which separates the entire robot trajectory into the global-action of approximate reaching to the target object and local-action of precise manipulation of the object, is used for the precise reaching subtask. (c) The output foveated vision of the proposed gaze-based visual attention can effectively eliminates the visual gap between the master and the robot.}
\label{fig:sample}
\end{figure*}

\IEEEPARstart{D}{eep} imitation learning is a model-free method for policy optimization that imitates expert (typically, human)-demonstrated behaviors using a deep neural network. This method has been applied to dexterous robot manipulation tasks for which the manual definition of the optimal solution is infeasible (e.g., \cite{kim2021gaze,kim2021transformer,zhang2018deep}).

\begin{table*}[]
\centering
\begin{tabular}{lcccc p{10cm}}
\hlineB{2}
Method                  & Safety & No visual distraction & Force feedback & No robot required during training            \\ \hline \hline
Master-to-robot (proposed)          & \greencheck & \greencheck            & \greencheck  & \greencheck\\
Teleoperation           & \greencheck & \greencheck         & \redx     & \redx\\
Bilateral teleoperation & \redx & \greencheck            & \greencheck  & \redx\\
Kinesthetic teaching    & \greencheck & \redx                & \greencheck  &\redx \\
Learning from watching  & \greencheck & \greencheck            & \redx  &\greencheck \\
\hlineB{2}
\end{tabular}
\caption{Comparison of demonstration methods.}
\label{tab:methods_comp}

\end{table*}

The aim of this study is the imitation learning of a manipulation task that requires force feedback for successful manipulation. 
Previous demonstration methods (i.e., teleoperation, bilateral teleoperation, kinesthetic teaching, and learning from watching) suffer when they consider force feedback during deep imitation learning (Table \ref{tab:methods_comp}).
In \textbf{teleoperation}-based methods in which the human operator teleoperates the robot using a controller (e.g., \cite{zhang2018deep,kim2020using}), force feedback is not provided to the human operator. \textbf{Bilateral} teleoperation enables accurate force control by reproducing the force feedback of the robot at the master's actuators (e.g., 
\cite{hulin2011dlr})
. However, in this method, the safety of the human operator who physically interacts with the moving actuators must be considered \cite{hulin2011dlr}. A complex control scheme is required to compensate for the time delay caused by communication between the master and robot \cite{guo2019scaled} or to ensure stability, and the system is expensive because both the master and robot require a motor. 
Previous approach to bilateral teleoperation with force feedback using a relatively low-cost and safe haptic device (Geomagic touch) was presented \cite{adachi2018imitation}, however, it lacks the power required for practical manipulation tasks.
\textbf{Kinesthetic teaching} is another robot teaching method in which a human grasps the robot arm and teaches the task by moving it (e.g., 
\cite{schou2013human,akgun2012trajectories}). 
However, when adapting the method to the end-to-end learning of visuomotor control, this method suffers from severe visual distractions, such as the human body. 
Additionally, this method requires a torque sensor on each joint for gravity and friction compensation; therefore, a high-cost force-controllable robot is required. 
A few researchers have studied a robot that \textbf{learns from watching} humans demonstrate tasks 
\cite{Kuniyoshi1994learning,yu2018one,liu2018imitation}. 
However, these methods cannot transfer force feedback because it cannot be acquired from video demonstrations. 
To summarize, current demonstration generation methods lack force feedback or require expensive robots in the demonstration system. This problem raises the demand for a simple demonstration method that can be applied to force feedback-based tasks. 

In this study, a simple master-only demonstration system that can imitate tasks that require force feedback (Fig. \ref{fig:sample}) is proposed. The proposed method includes the hardware structure for the demonstration, methods to reduce the domain gap between the demonstration and the real-robot environment, and dual-action-based deep imitation learning with Transformer-based force/torque (F/T) sensory attention.

A master controller is proposed that imitates the same Denavit-Hartenberg (DH) parameters as the UR5 (Universal Robots) robot, and the robot and controller have the same end-effector design and F/T sensors (Fig. \ref{fig:gp_master} and \ref{fig:gp_robot}). The human demonstrator uses this controller to directly manipulate the object and receive force feedback during manipulation. The shared DH parameters between the master and robot facilitate the transfer of motor skills from humans to robots by constraining the human operators' physicality through the use of the master controller.
This master controller for the M2R policy transfer system has a few advantages. First, a demonstration with force feedback is possible without an expensive bilateral system; this controller does not require any motor or gear, only a simple encoder. 
Second, no complex control scheme is required to ensure safety and stability because the human controls the end-effector while performing the demonstration.
Third, this controller can teach tasks that require force feedback with the Transformer-based F/T sensory attention.

This system must overcome two types of domain gaps between the master and robot: visual gap and kinematic gap. 
The \textbf{visual gap} is a domain gap in vision caused by different visual components, such as the human arm. 
Gaze-based deep imitation learning \cite{kim2020using} has been proposed, which can suppress the visual gap using the human gaze that is naturally estimated during the demonstration. Because the human naturally gazes at the target object \cite{hayhoe2005eye}, foveated vision cropped from global vision using the predicted gaze position effectively suppresses the visual gap.

The \textbf{kinematic gap} is a domain gap between the master controller and robot caused by their kinematic difference. This study uses a master controller with the same DH parameters as the robot to minimize the kinematic gap.
Additionally, a simple calibration method is used to minimize the kinematic difference between the master and robot caused by deflection. 

In this study, the Transformer architecture  \cite{vaswani2017attention} is used to process force/torque (F/T) sensory information to make the robot attend to important channels from the relatively high-dimensional dual-arm F/T sensory inputs. Additionally, dual-action separation\cite{kim2021gaze} is used to ensure precise object manipulation.

The proposed imitation learning method, accompanied by a master-only demonstration system, was validated in the real-world robot experiment of a bottle-cap-opening task, which requires both a precise grasping-cap policy, and the policy considers force feedback.

To summarize, the contributions of this study are as follows:
\begin{itemize}
    \item Master-to-robot policy transfer system is proposed that can learn the task requires force feedback with the master-only demonstration data.
    \item Gaze-based visual attention is proposed that can minimize the visual gap.
    \item Transformer-based F/T sensory input attention is proposed that can train a robot on tasks that require force feedback.
\end{itemize}

\section{Method}
The proposed M2R policy transfer system consists of a hardware for master-only demonstrations, gaze-based visual attention and a kinematic calibration method to minimize the domain gap between the master and robot, and dual-action deep imitation learning with the Transformer-based F/T sensory attention. In this section, each component is described.

\subsection{Hardware for master-only demonstrations}
\begin{figure}
  \centering
  \vspace{0.0in}
  \begin{subfigure}[t]{0.4\linewidth}
    \centering
    \captionsetup{width=1.\linewidth}
    \captionsetup{justification=centering}
    \includegraphics[width=1.\linewidth]{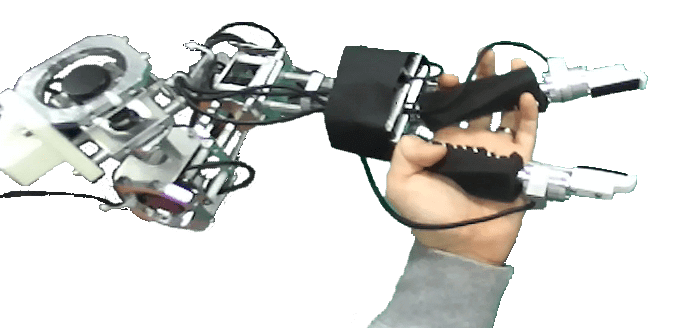}
    \caption{Master controller and end-effector.}
    \label{fig:gp_master}
  \end{subfigure}%
  \hfil
  \begin{subfigure}[t]{0.4\linewidth}
    \centering
    \captionsetup{width=1.\linewidth}
    \captionsetup{justification=centering}
    \includegraphics[width=1.\linewidth]{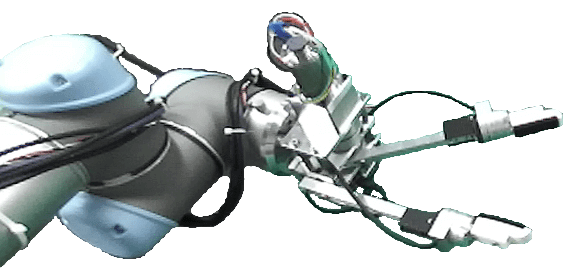}
    \caption{UR5 robot and end-effector.}
    \label{fig:gp_robot}
  \end{subfigure}%
  \hfil
  \begin{subfigure}[t]{0.5\linewidth}
    \centering
    \captionsetup{width=.8\linewidth}
    \captionsetup{justification=centering}
    \includegraphics[width=0.8\linewidth]{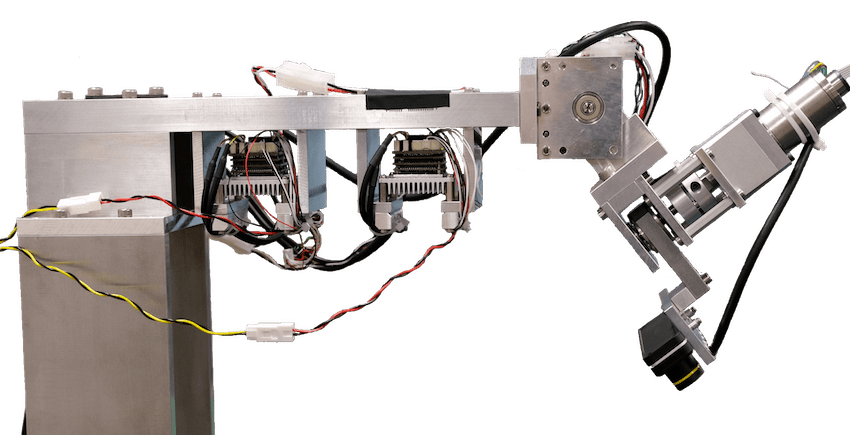}
    \caption{Camera mount.}
    \label{fig:camera}
  \end{subfigure}%

  \caption{Hardware components of the M2R policy transfer system. The master controller (\ref{fig:gp_master}) and robot (\ref{fig:gp_robot}) share the same DH parameter and the same type of fingertip. (\ref{fig:camera}) Camera mount is attached to both the master and robot.}
  \label{fig:gp}
 \end{figure}

This framework uses a dual-arm robot system with two UR5 robots. In the authors' previous study (e.g., \cite{kim2020using}), this system was operated in teleoperation mode. By contrast, in the present study, demonstration data are generated in master-only mode. In this mode, the human demonstrator uses the fingertip of the master controller (Fig. \ref{fig:gp_master}) to execute tasks. The controller collects joint angles from its encoders mounted on a structure that has DH parameters identical to those of the UR5 robot. 
The shared DH parameters between the master and robot facilitate the transfer of motor skills from humans to robots. The idea behind this is that the physicality of the robot has a significant influence on motor skills, and constraining the physicality of the human operators through the use of the master controller allows them to develop motor skills suitable for the robot. This helps to ensure that the kinematics of the system can be solved. 
Therefore, the robot can easily reproduce human behavior by following the recorded encoder values. The end effectors on the robot and master are identical: one degree-of-freedom grippers (Fig. \ref{fig:gp_robot}). 
Additionally, an F/T sensor (Leptrino, PFS 030YA301) is placed between the fingertip and rod so that the F/T of the fingertip can be measured.

The human operator observes a stereo image from a stereo camera (ZED-Mini, StereoLabs) using a head-mounted display (HMD). The stereo cameras are mounted in identical relational positions on both the robot and master sides (Fig. \ref{fig:camera}). For the master, this location is just above the operator's head. This shared camera mount position minimizes the visual gap between the master and robot caused by different viewpoints. The developed pan-tilt camera system can control its direction, which can be used in future studies.

\subsection{Calibration}
The proposed controller and UR5 robot have identical DH parameters. However, the authors found that there is still an error because of deflection caused by gravity. Therefore, calibration is required to reduce the error.

Motion capture is used for calibration. The accurate end effector position and rotation are measured using the Optitrack motion capture device. The master controller is moved randomly by the human operator while it records the joint angle, and the motion capture device measures the exact position simultaneously, which is then played back in the robot.
A transformation from the motion-captured master position/rotation $M^{mocap}$ to the robot position/rotation $R^{mocap}$ transforms the position/rotation $M^{enc}$ calculated from the joint angle into the calibrated robot state $R^{enc}$. 

The position $p$ and orientation $o$ are separated, and the homogeneous transformation matrix $A_p$ and rotation matrix $A_o$ are calculated by learning the least-squares solution: 
\begin{equation}
\begin{aligned}
    R_p^{mocap} = A_p^{4 \times 4} M_p^{mocap}
\end{aligned}
\end{equation}
\begin{equation}
\begin{aligned}
    R_o^{mocap} = A_o^{3 \times 3} M_o^{mocap}.
\end{aligned}
\end{equation}

Therefore, the calibration of robot state $R^{enc}$ is calculated using the following equations:
\begin{equation}
\begin{aligned}
    R_p^{enc} = A_p^{4 \times 4} M_p^{enc}
    \end{aligned}
\end{equation}
\begin{equation}
\begin{aligned}
    R_o^{enc} = A_o^{3 \times 3} M_o^{enc},
    \end{aligned}
\end{equation}
and the calibrated robot state $R^{enc}$ is used for neural network model training.

\subsection{Gaze-based visual attention}

Gaze-based visual attention was proposed in \cite{kim2020using}. This method reproduces the human's gaze, which can be acquired naturally during demonstration using an eye tracker on the HMD (Fig. \ref{fig:sample}). The gaze-based visual attention is used to minimize the visual gap between the demonstration data from the master and the test on the robot. Because the human gaze is correlated with the target object position \cite{hayhoe2005eye,pelz2001coordination}, this method can eliminate distractions from the human hand. Therefore, the visual appearance of both the robot and master are similar in foveated vision  (Fig. \ref{fig:sample} (c)).

This method first collects the two-dimensional gaze coordinates using the eye tracker during the demonstration. Then, mixture density network (MDN)-based \cite{bishop1994mixture} gaze prediction architecture \cite{bazzani2016recurrent,kim2020using} is trained to reproduce the probability of the gaze position based on the global vision. The trained MDN predicts the gaze position in the test environment using robots. Foveated vision is an area around the gaze position cropped from the global vision. This visual attention process can suppress visual distractions from task-unrelated objects or background while fully providing information about the task-related object or area.

The details of MDN architecture are as follows: First, a series of five convolutional neural networks (CNNs) and rectified linear units (ReLUs) followed by SpatialSoftmax \cite{finn2016deep} extract visual features, which are then processed by the MDN, which is composed of a Gaussian mixture model (GMM) with eight Gaussians. Finally, this GMM is fit to the coordinates of the gaze using \cite{kim2021transformer}
\begin{equation}
\label{eq:loss_gaze}
\mathcal{L}_{gaze} = -log \big{(}\sum_{i=1}^8 p^i \mathcal{N} (g;\mu^i,\mathbf{\Sigma}^i)\big{),}
\end{equation}
where $p^i, \mu^i,$ and $\mathbf{\Sigma}^i$ represent the weight, mean, and covariance matrix of the $i$th Gaussian, respectively, and $g$ denotes a target gaze coordinate. 
The gaze position is used to crop $128 \times 128$ foveated vision from $256 \times 256$ global vision.

It is important to note that kinesthetic teaching is unable to utilize the advantage of gaze-based attention as the human operator must control the dual-arm robot from the opposite side of the table, which creates confusion when attempting to observe the view from the robot's camera through the HMD.

\subsection{Dual-action}
\begin{figure*}
\centering 
\vspace{0.1in}
\includegraphics[width=.70\linewidth]{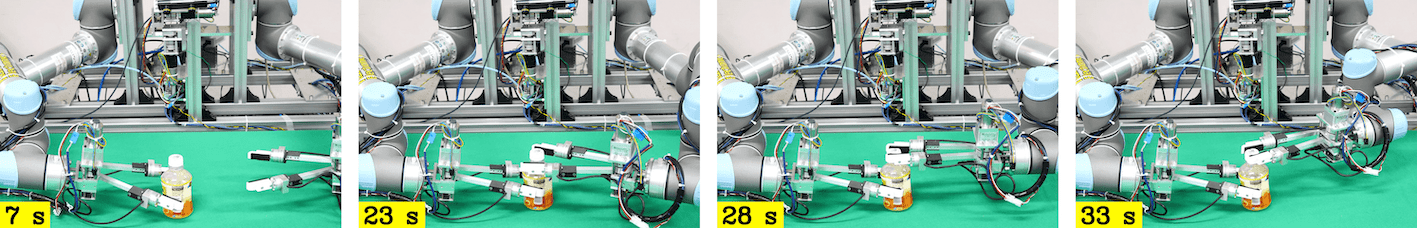}
\caption{Example of the successful behavior of the proposed \textit{DA-force}.
\textit{DA-force} successfully grasped the bottle ($\sim 7s$), grasped the cap ($7s \sim 28s$), and rotated it ($28s \sim 33s$).}
\label{fig:success_example}
\end{figure*}

Because cap-reaching behavior requires precise manipulation, the dual-action architecture proposed in \cite{kim2021gaze} is used. Based on a physiological study \cite{paillard1996fast} in which the human's reaching movement was divided into two distinctive systems, the dual-action architecture divides the entire robot trajectory into the global action and local action. This architecture first detects whether the end effector is in foveated vision \cite{kim2022robot}: if not, the global action, which is the fast action that delivers the hand to the vicinity of the target, is used; and if so, the local action, which is a precise and slow action when the hand is near the target, is activated (Fig. \ref{fig:sample}b).
This enables high-precision manipulation, such as needle threading \cite{kim2021gaze} or banana peeling \cite{kim2022robot}.

In the dual-action architecture, each step is labeled as \textit{global-action}, \textit{local-action} from the demonstration data and used to train the global-action network or local-action network, respectively. A human manually annotated some demonstrations to train a CNN-based classifier to classify the remaining demonstrations. The classifier's output was verified by a human (The details can be found at \ref{sec:demo_gen}). 
The global and local-action networks are distinctively trained using the annotated data. During the test, a binary classifier, which inputs foveated vision to classify the global and local actions, is used to choose which action network to use.

\begin{figure*}
  \centering
  \vspace{0.0in}
  \begin{subfigure}[t]{0.35\linewidth}
    \centering
    \captionsetup{width=.8\linewidth}
    \captionsetup{justification=centering}
    \includegraphics[height=0.20\textheight]{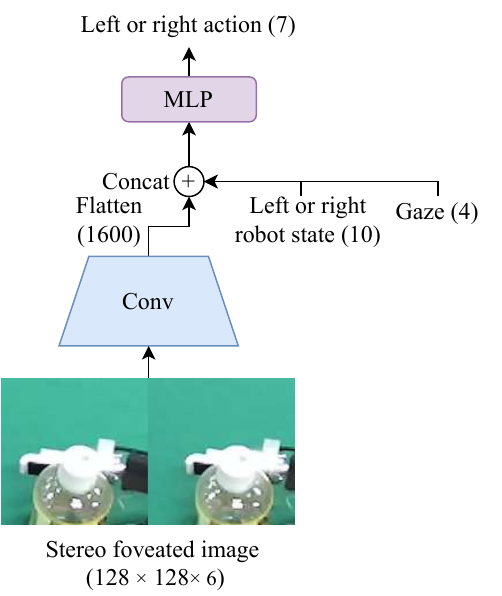}
    \caption{Global-action network}
    \label{fig:global_network}
  \end{subfigure}%
  \begin{subfigure}[t]{0.25\linewidth}
    \centering
    \captionsetup{width=.8\linewidth}
    \captionsetup{justification=centering}
    \includegraphics[height=0.20\textheight]{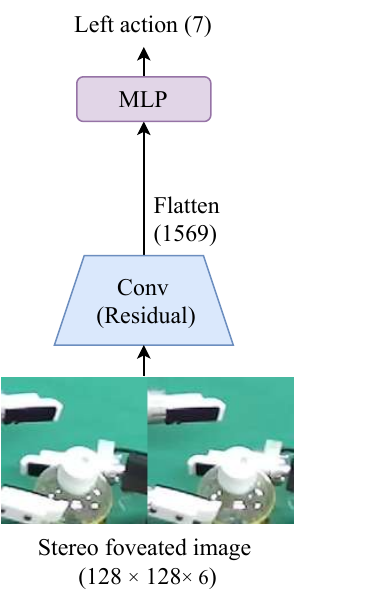}
    \caption{Local-action network}
    \label{fig:local_network}
  \end{subfigure}%
  \begin{subfigure}[t]{0.45\linewidth}
    \centering
    \captionsetup{width=.8\linewidth}
    \captionsetup{justification=centering}
    \includegraphics[height=0.20\textheight]{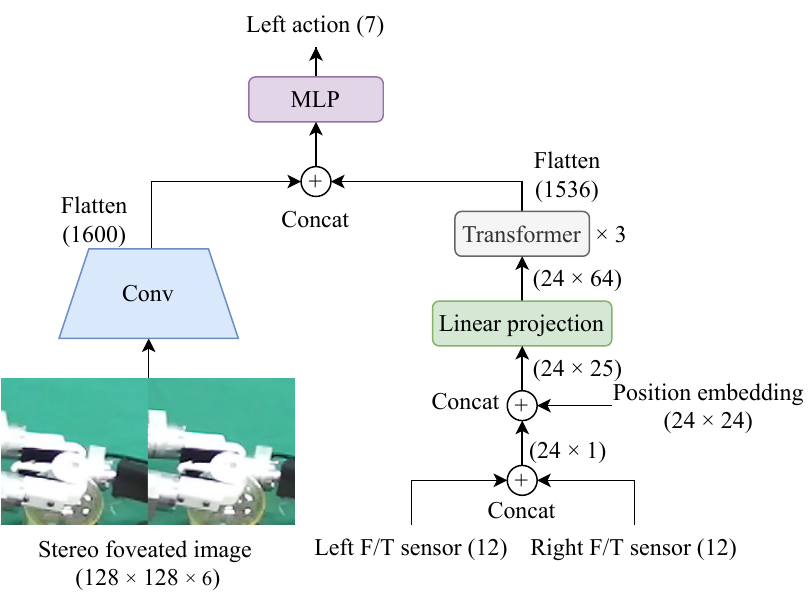}
    \caption{Force feedback network}
    \label{fig:force_feedback_network}
  \end{subfigure}%

  \captionsetup{justification=centering}
  \caption{Network architectures; the number in brackets indicates the dimensions of the data.}
  \label{fig:network}
 \end{figure*}

The global-action network inputs stereo foveated images, the robot state, and the stereo gaze position to output the action (Fig. \ref{fig:global_network}). 
The local-action network inputs foveated vision to output the precise local action. Because the robot's kinematic state is not used for input, this network can accurately predict the action regardless of the kinematic gap between the demonstration data from the master and the test environment of the robot (Fig. \ref{fig:local_network}).

The action is defined as the difference between the seven-dimensional robot state (three-dimensional position, three-dimensional orientation, and a gripper angle) at the next step and the current robot state. The orientations of the robot state input into the neural network models are decomposed using cosine and sine to prevent the drastic change of the angle representation; therefore, it is ten-dimensional. The global-action network uses a series of CNN, ReLU, and max-pooling layers with the stride equal to 2. 
The local-action network uses one CNN layer with an ReLU followed by nine residual blocks \cite{he2016deep}, and a max-pooling layer is inserted in every two residual blocks with the stride equal to 2. The CNN layer uses a $3 \times 3$ filter with 32 channels. 
The multilayer perceptron (MLP) module is a series of fully connected layers (FCs) with a node size of 200 and an ReLU between the FCs. The action is optimized using $\ell_2$ loss.

\subsection{Transformer-based force-feedback network}
The force feedback network uses the dual-arm F/T sensory data and foveated image to compute the action command (Fig. \ref{fig:network}). This architecture processes F/T sensory data with the Transformer-based self-attention architecture \cite{vaswani2017attention}, inspired by a study \cite{kim2021transformer} in which researchers implemented the self-attention architecture to reduce distractions in dual-arm somatosensory inputs. In the present study, the Transformer is used to attend to useful F/T sensory information from both arms and reduce distractions from other F/T sensory information. The feed-forward network also processes the foveated image using the same CNN architecture as the global-action network. The output of the CNN and Transformer is concatenated and then processed with the MLP to predict the action command.

\begin{table*}[]
\centering
\caption{Subtask description}
\begin{tabular}{lcccc}
\hlineB{2}
Subtask     & Init & Goal & Network type   & Description                                        \\\hline \hline
\textit{GraspCap}    & \adjustbox{valign=c}{\includegraphics[width=3cm]{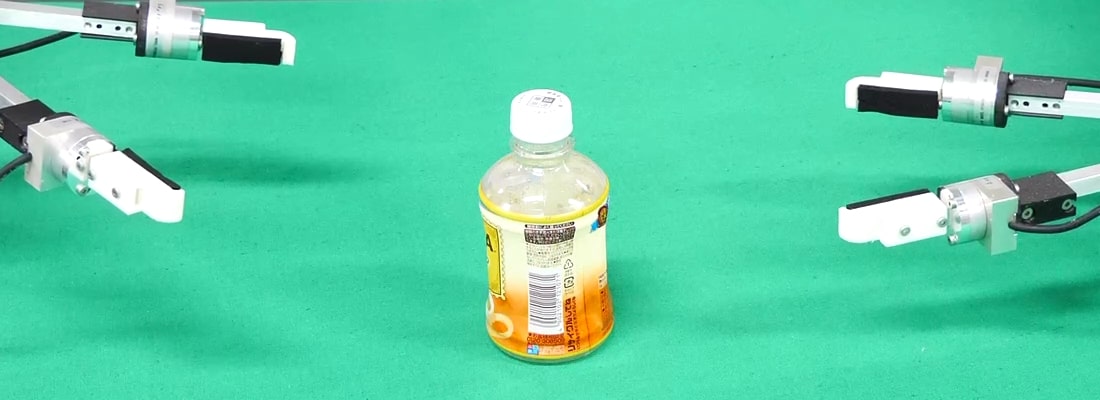}}     & \adjustbox{valign=c}{\includegraphics[width=3cm]{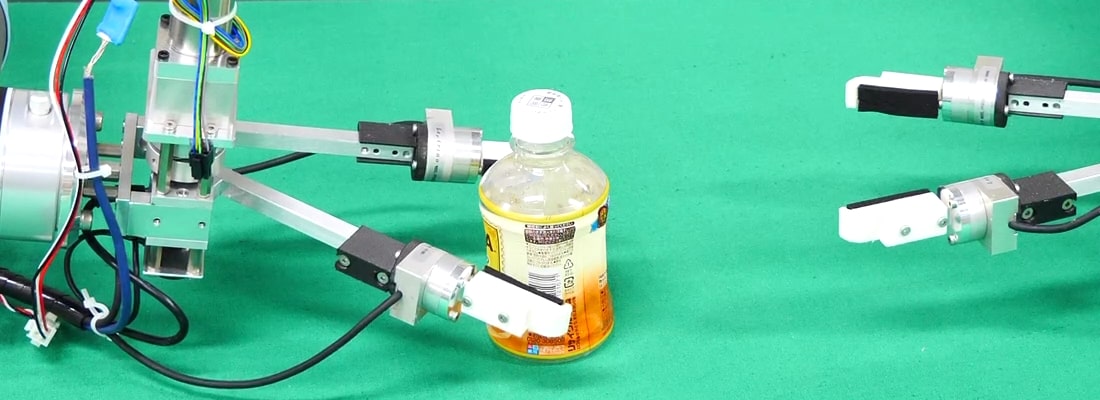}}     & Global         & Grasp the body of the bottle using the right hand \\\hline
\textit{GraspBottle} & \adjustbox{valign=c}{\includegraphics[width=3cm]{task_1_crop.jpg}}     & \adjustbox{valign=c}{\includegraphics[width=3cm]{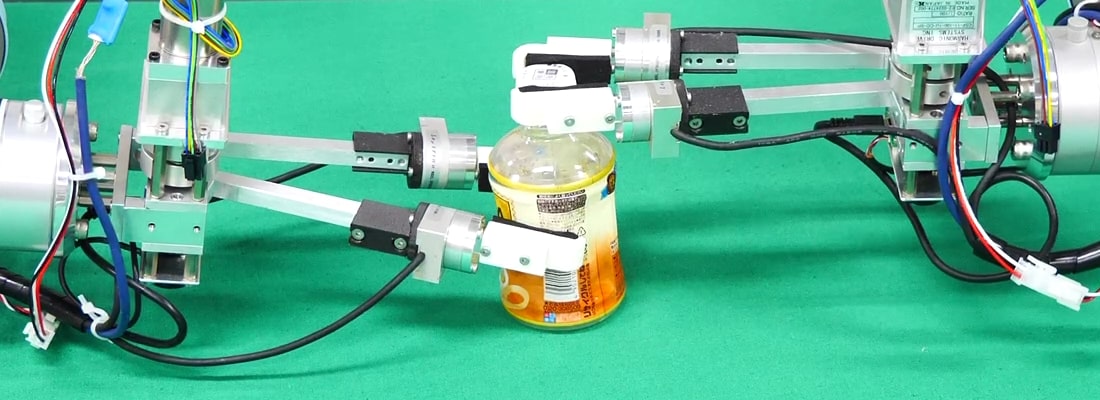}}     & Global $\rightarrow$ Local   & Reach and grasp the bottle cap with the left hand \\\hline
\textit{Rotate}      & \adjustbox{valign=c}{\includegraphics[width=3cm]{task_3_crop.jpg}}    & \adjustbox{valign=c}{\includegraphics[width=3cm]{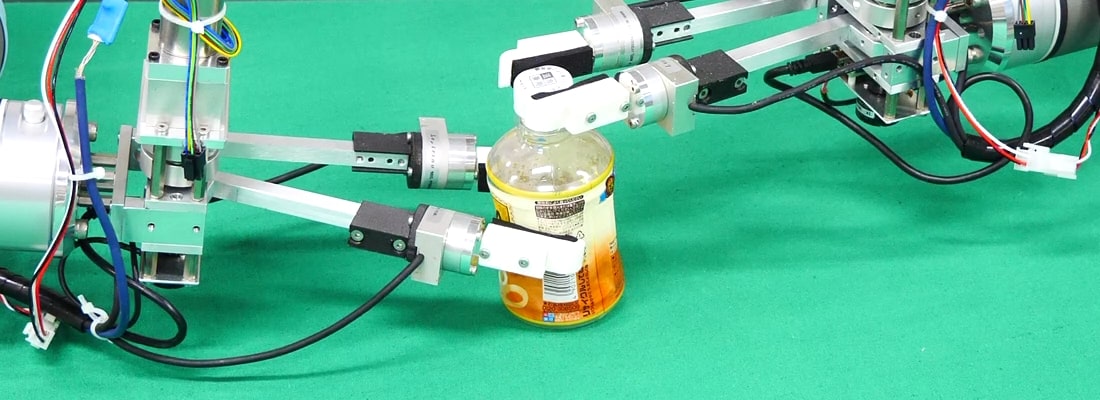}}    & Force feedback & Open the bottle cap by rotating the left hand   \\
\hlineB{2} 
\end{tabular}
\label{tab:task_description}
\end{table*}

\section{Experiments}
\subsection{Calibration results}

\begin{figure}
  \centering
  \vspace{0.0in}
  \begin{subfigure}[t]{.5\linewidth}
    \centering
    \captionsetup{width=.95\linewidth}
    \includegraphics[width=0.95\linewidth]{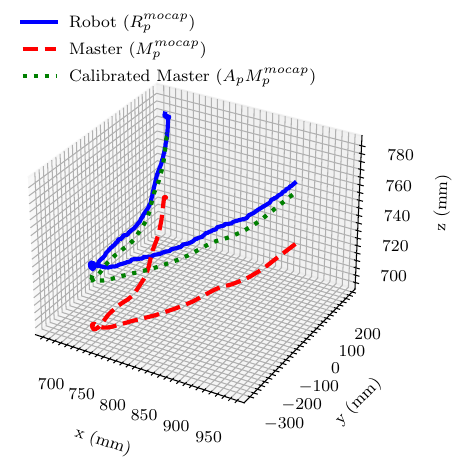}
    \caption{Visualized trajectory of the robot’s left arm. Simple linear calibration can transform the master's trajectory into the robot's trajectory.}
    \label{fig:calib_example}
  \end{subfigure}%
  \hfil
  \begin{subfigure}[t]{0.5\linewidth}
    \centering

    \captionsetup{width=.95\linewidth}
    \includegraphics[width=0.95\linewidth]{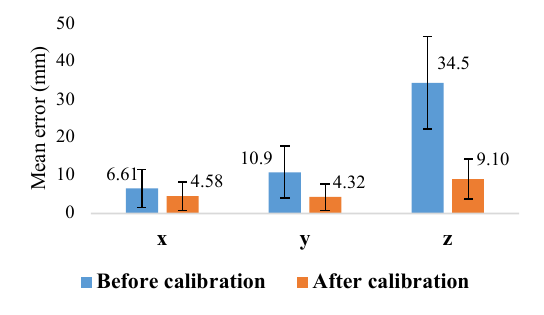}
    \caption{Calibration reduces the kinematic error between the master and robot.}
    \label{fig:calib_error}
  \end{subfigure}%

  \captionsetup{justification=centering}
  \caption{Calibration result.}
  \label{fig:calib}
 \end{figure}

Fig. \ref{fig:calib_example} illustrates part of the calibrated trajectory of the left arm. After calibration, the master controller's position $M_p^{mocap}$ is translated into $A_p M_p^{mocap}$, which is much closer to the actual robot position $R_p^{mocap}$. Fig. \ref{fig:calib_error} compares the mean Euclidean distance between the master and robot,  $||M_p^{mocap} -R_p^{mocap}||$ (blue), and after calibration, $||A_p M_p^{mocap} - R_p^{mocap}||$ (orange). The result indicates that the overall error decreased after calibration. Particularly, the error on the $z$-axis, which was mainly caused by the deflection of the upper arm because of gravity, occupied the most significant error between the master and robot, but decreased after calibration. This calibration was robust because the main error on the $z$-axis was caused by gravity, which is almost constant on the surface of the earth.

\subsection{Bottle-cap-opening task}
Our previous robot system successfully solved various dexterous manipulation tasks (e.g., \cite{kim2021gaze,kim2021transformer,kim2022robot}) by demonstration with teleoperation. However, we encountered limitations in achieving contact-rich tasks with strict geometrical constraints, which even human teleoperators could not accomplish. The proposed M2R framework targets bottle-opening, a representative example of such tasks that requires precise gripper rotation with exact ($x$, $y$) translation to compensate for the offset between the cap and gripper. The M2R framework enables the demonstrator to directly feel force-feedback from the exerted action, facilitating the discovery of motor skills suitable for the robot in contact-rich tasks.

\subsection{Training setup}
\begin{figure}
  \centering
  \vspace{0.0in}
  \includegraphics[width=0.5\linewidth]{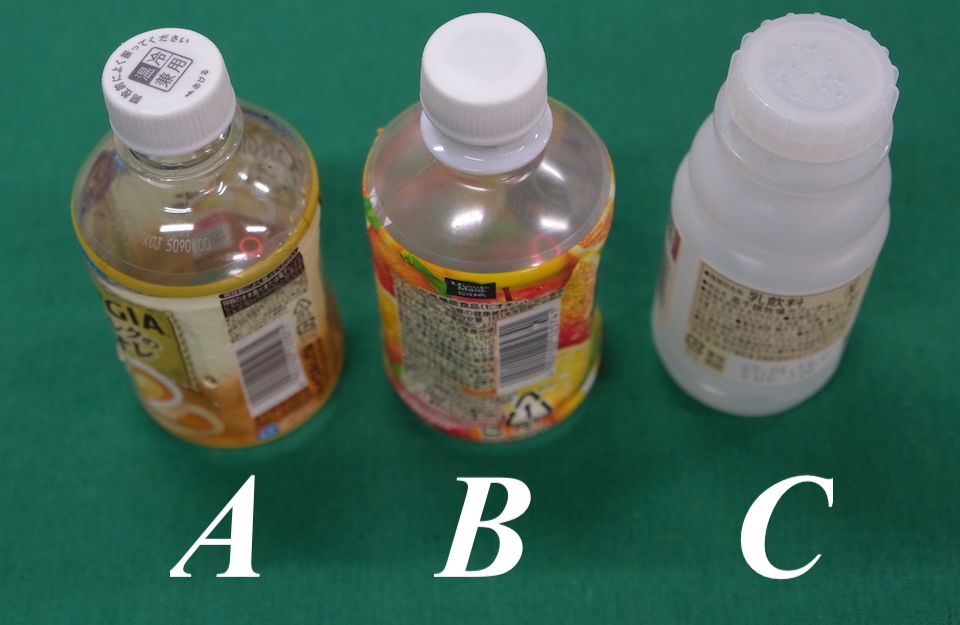}
  \captionsetup{justification=centering}
  \caption{Bottles used in experiment.}
  \label{fig:bottles}
 \end{figure}

We trained and tested our model with three different plastic bottles (\textit{bottle-A,B,} and \textit{C}, Fig.  \ref{fig:bottles}) We used \textit{bottle-A} for ablation studies and other analysis. Relatively small number of  demonstration sets of \textit{bottle-B} and \textit{C} (Table \ref{tab:dataset_statistics} ) are added to and re-trained to evaluate the adaptation ability of the proposed model on new bottle with less dataset. \textit{Bottle-B} has the same size and shape but different texture with the \textit{bottle-A}. \textit{Bottle-C} has the different size, shape and texture, thus require more adaptation ability than \textit{bottle-B}.

The bottle-cap-opening task was segmented into the three subtasks \textit{GraspBottle}, \textit{GraspCap}, and \textit{Rotate} described in Table \ref{tab:task_description}. \textit{GraspBottle} $\rightarrow$ \textit{GraspCap} was segmented because the robot changes its manipulating arm from right to left, and then \textit{GraspCap} $\rightarrow$ \textit{Rotate} because rotation requires force feedback, whereas \textit{GraspCap} does not. Each subtask used a different neural-network architecture. First, \textit{GraspBottle} is a simple grasping movement that the global-action network can accomplish. \textit{GraspCap} is high-precision grasping, which requires a dual-action network that combines global and local-action networks. Finally, \textit{Rotate} requires force feedback; therefore, a force feedback network was used.

\subsubsection{Demonstration data generation} \label{sec:demo_gen}
During the demonstration, the human operator grasped the bottle with the right gripper (\textit{GraspBottle}), reached the bottle cap with the left gripper (\textit{GraspCap}), and repeatedly rotated the cap until it opened (\textit{Rotate}). Each subtask was segmented during training using a foot keyboard. \textit{GraspCap} was more complicated than the other subtasks because the exact grasping of the cap requires reaching to the cap with small clearance (i.e., small mistake in grasping results in failure to open the cap during \textit{Rotate}).
Therefore, demonstrations that repeated \textit{GraspCap} were added. The entire demonstration set was divided into $90\%$ training and $10\%$ validation sets. The training dataset statistics are presented in Table \ref{tab:dataset_statistics}.

\begin{table}
\centering
\begin{tabular}{lccc}
\hlineB{2}
    Dataset  & Bottle index & \# of demos & Total demo time (min) \\ \hline \hline
\textit{GraspBottle}   & \textit{A} &  312 & 9.475 \\
                    & \textit{B} &  297 & 8.192 \\
                    & \textit{C} &  442 & 10.49 \\

\textit{GraspCap}     & \textit{A} &  5798 & 156.0 \\
                    & \textit{B} &  1029 & 25.81 \\
                    & \textit{C} &  1063 & 29.30 \\
\textit{Rotate}     & \textit{A} &  1111 & 23.75 \\
                    & \textit{B} &  329 & 5.547 \\
                    & \textit{C} &  329 & 5.025 \\
\hlineB{2}

\end{tabular}
\caption{Training dataset statistics.}
\label{tab:dataset_statistics}
\end{table}

From a $1280 \times 720 \times 3$ raw RGB stereo camera image, a region of $256 \times 256$ pixels was cropped to form the global vision around the center position of $[678, 428]$ for left vision and $[756, 428]$ for right vision to reduce the visual gap.

The global/local-action labels were annotated in \textit{GraspCap}. First, a human manually annotated $202$ demonstrations from dataset of \textit{bottle-A}. The result labels were used to train a simple CNN-based binary classifier $\pi(f_t) \rightarrow \{\textit{global-action}, \textit{local-action}\}$, where $f_t$ denotes the foveated image. The trained classifier then automatically classified the remaining 5596 demonstrations. If the classification result of one demonstration episode $f_0, ..., f_t$ provided more than one transition (i.e., there was any $t$ of which $\pi(f_t-1) \rightarrow \textit{local-action}$ and $\pi(f_t) \rightarrow \textit{global-action}$ for any time step $t \in [1, L]$), the human reannotated the demonstration episode. Datasets of \textit{bottle-B} and \textit{bottle-C} were annotated with same process.

\subsubsection{Model training}
The gaze predictor and policy network were trained on each subtask. These models were trained for 300 epochs with a learning rate of $3e-5$ using rectified Adam \cite{liu2019variance} and a weight decay of 0.01. For training, eight NVIDIA v100 GPUs with Intel Xeon CPU E5-2698 v4 or four NVIDIA a100 GPUs with two AMD EPYC 7402 CPUs were used, and for the robot tests, one NVIDIA GTX 1080 and one Intel i7-8700K CPU were used.

\subsection{Test setup}\label{sec:test_setup}
The objective of the test was to evaluate whether the robot could grasp the bottle with its right hand, grasp the bottle cap with its left hand, and rotate the cap once without the cap slipping or bottle tilting. The test was repeated 18 times for different initial bottle positions, which were reproduced from the initial positions recorded in the randomly sampled validation set. The bottle cap was manually rotated approximately 90 degrees from the completed closed state because opening the completely locked cap required too much torque for the robot.

This study proposed the gaze-based dual-action approach with force feedback. Four types of neural network architecture were compared to evaluate the validity of each component:
\begin{itemize}
    \item \textit{DA-force}: \textit{DA-force} uses both F/T sensory feedback and gaze-based dual-action (DA) architecture.
    \item \textit{No-force}: 
    to validate the necessity of an F/T sensor, \textit{No-force} does not input F/T sensor information while rotating the cap.
    \item \textit{No-DA}: this architecture validates whether dual action is effective in the M2R framework. In this architecture, global and local actions are not distinguished. Therefore, the policy in each subtask is trained and inferred only from the global-action network. 
    \item \textit{No-gaze}: 
    \textit{DA-force} inputs foveated images to minimize the visual gap. \textit{No-gaze} inputs global vision to investigate whether the visual gap is indeed reduced by the gaze. As proposed in \cite{zhang2018deep}, the global image was processed using SpatialSoftmax after five layers of CNNs and ReLU.
\end{itemize}

\subsection{Experimental results}

\begin{figure}
  \centering
  \vspace{0.0in}
  \includegraphics[width=0.9\linewidth]{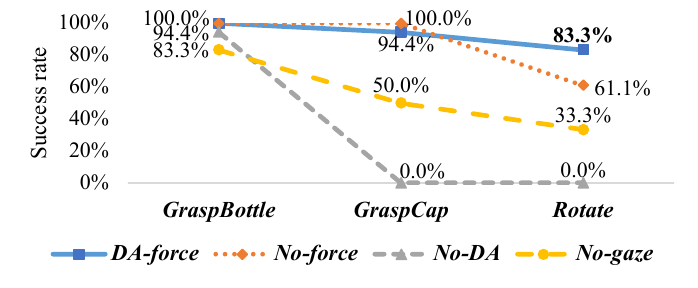}
  \captionsetup{justification=centering}
  \caption{Cumulative success rate (18 trials, \textit{bottle-A}}).
  \label{fig:success_rate}
 \end{figure}

\begin{table}
\centering
\begin{tabular}{lccc}
\hlineB{2}
    Bottle index  & \textit{GraspBottle} & \textit{GraspCap} & \textit{Rotate} \\ \hline \hline
\textit{A}     & 100\% &  94.4\% & 83.3\% \\
\textit{B}     & 100\% &  94.4\% & 77.8\% \\
\textit{C}     & 88.9\% &  77.8\% & 77.8\% \\
\hlineB{2}
\end{tabular}
\caption{Success rate comparison of all bottles.}
\label{tab:success_rate_bottles}
\end{table}

\begin{figure}
  \centering
  \vspace{0.0in}
  \begin{subfigure}[t]{0.5\linewidth}
    \centering
    \captionsetup{width=0.95\linewidth}
    \captionsetup{justification=centering}
    \includegraphics[width=0.95\linewidth]{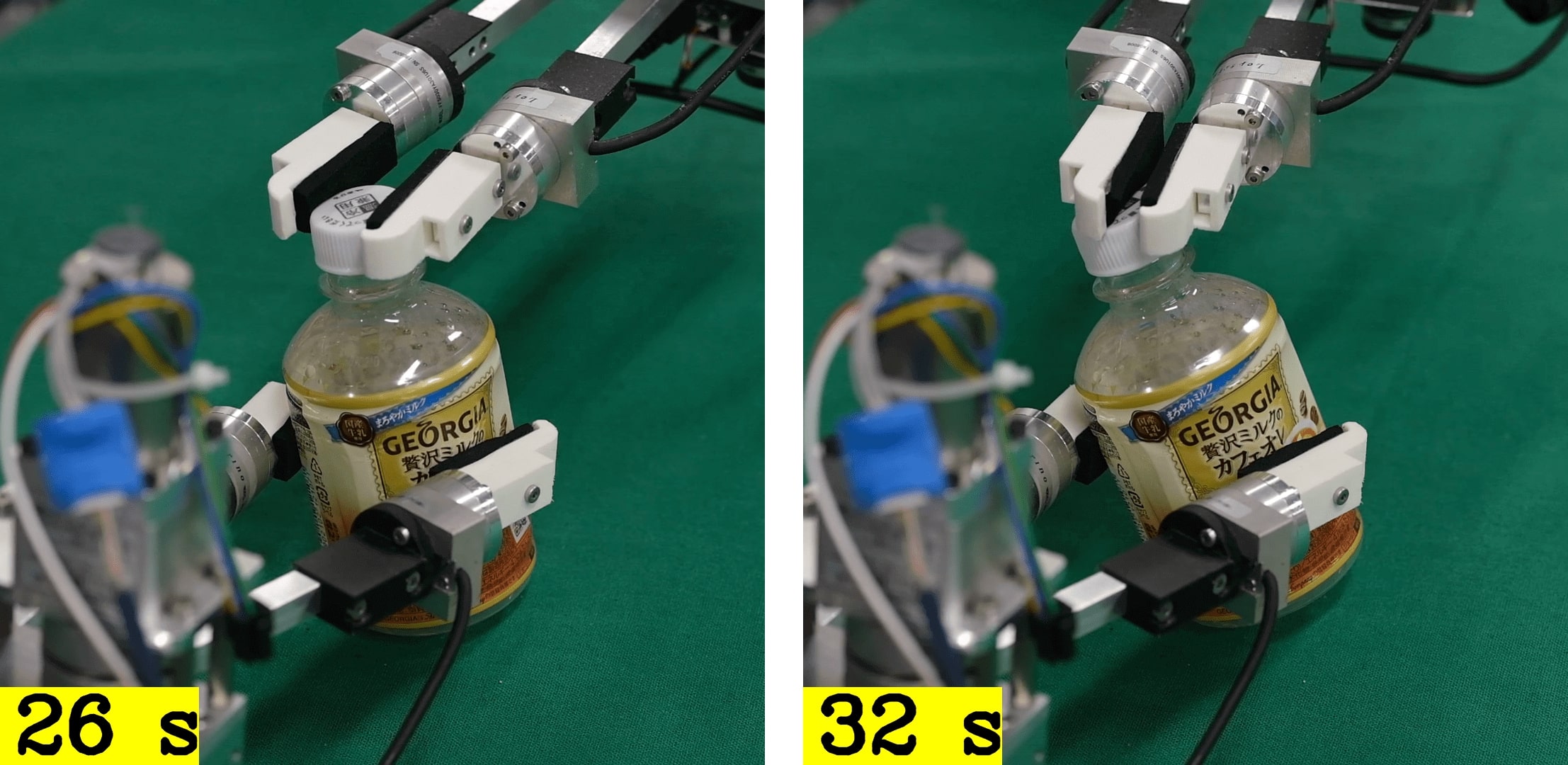}
    \caption{Failure example of \textit{No-force}: fingertip slipped on the cap during rotation.}
    \label{fig:failure_nosensor}
  \end{subfigure}%
  \begin{subfigure}[t]{.5\linewidth}
    \centering
    \captionsetup{width=0.95\linewidth}
    \captionsetup{justification=centering}
    \includegraphics[width=0.95\linewidth]{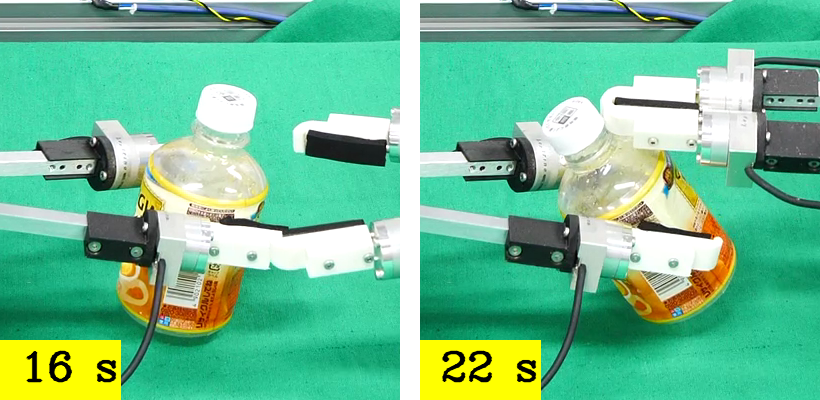}
    \caption{Failure example of \textit{No-gaze}: robot was not able to correctly grasp the bottle cap.}
    \label{fig:failure_global}
  \end{subfigure}%
  \hfil

  \begin{subfigure}[t]{1.0\linewidth}
    \centering
    \captionsetup{width=1\linewidth}
    \captionsetup{justification=centering}
    \includegraphics[width=1\linewidth]{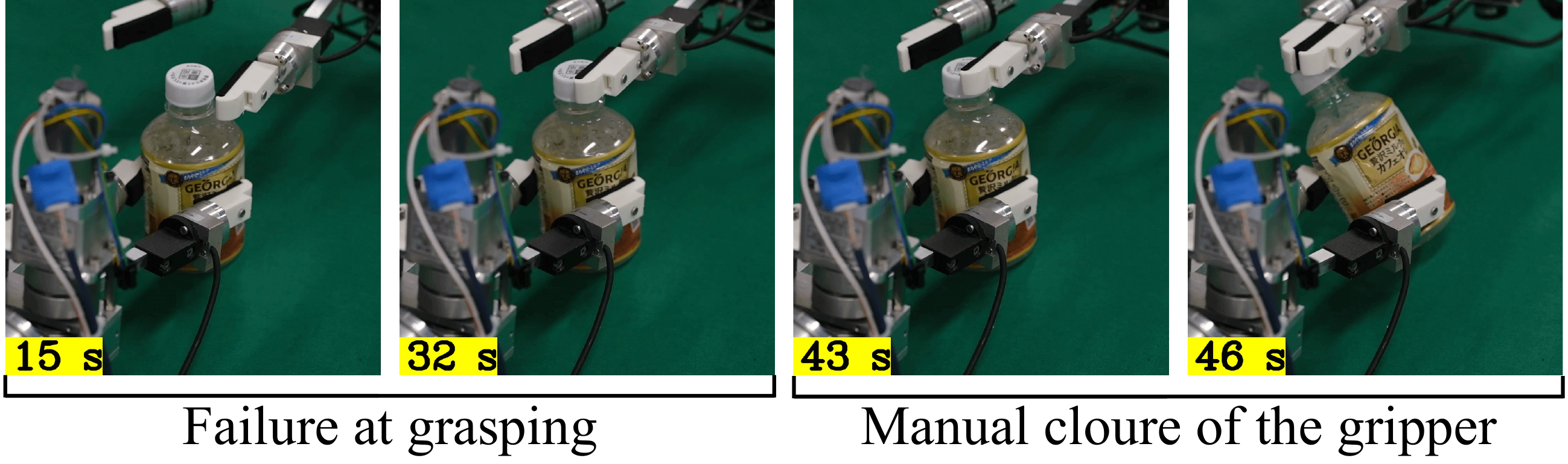}
    \caption{Failure example of \textit{No-DA}: robot failed to grasp the bottle cap. When the gripper was manually closed, the bottle collapsed.}
    \label{fig:failure_nodual}
  \end{subfigure}%
  \caption{Failure examples.}
  \label{fig:failures}
\end{figure}

\begin{figure}
  \centering
  \vspace{0.0in}
  \begin{subfigure}[t]{0.25\linewidth}
    \centering
    \captionsetup{width=.9\linewidth}
    \captionsetup{justification=centering}
    \includegraphics[width=0.9\linewidth]{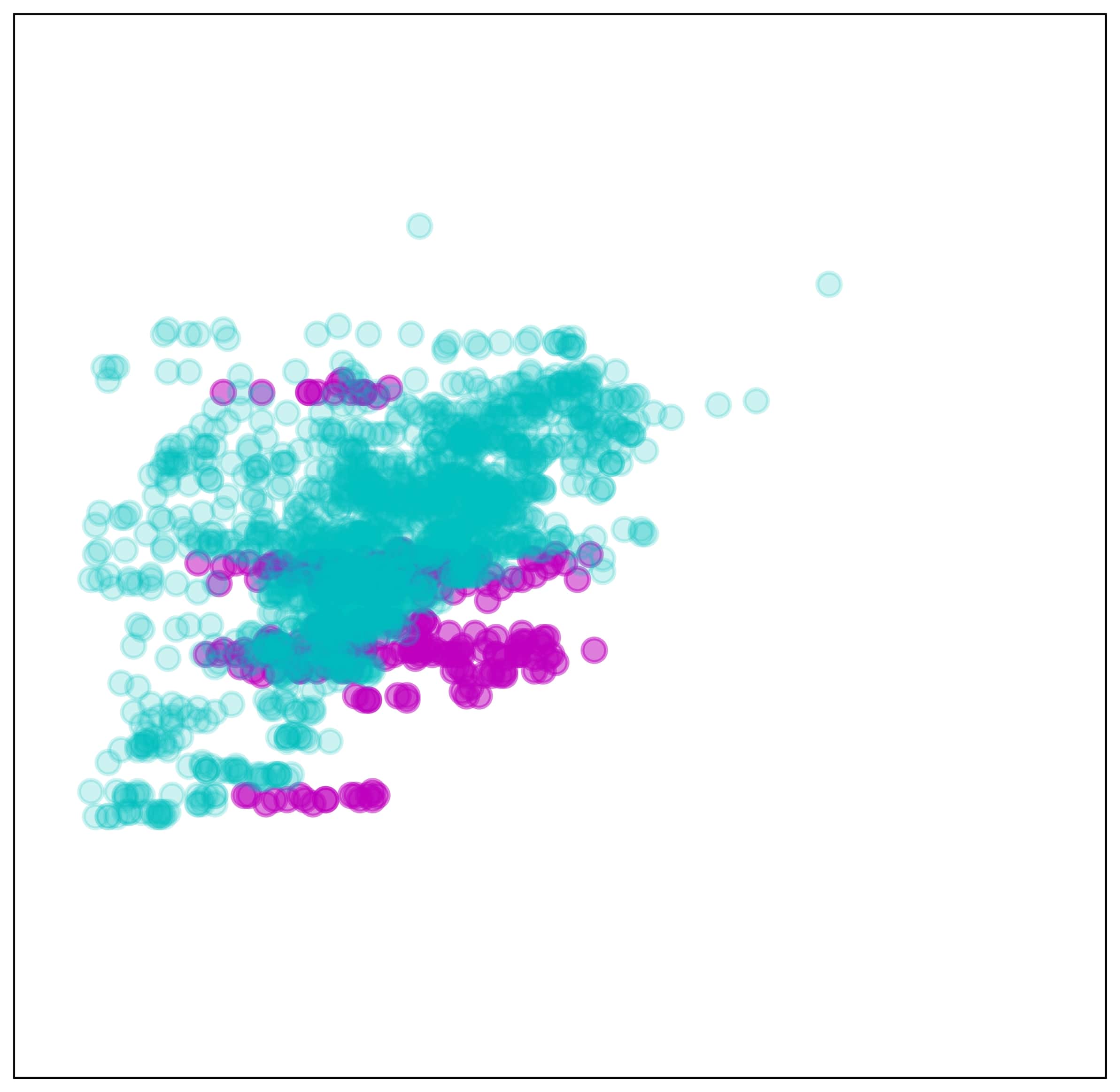}
    \caption{$16$.}
    \label{fig:ssm_16}
  \end{subfigure}%
  \begin{subfigure}[t]{0.25\linewidth}
      \centering
    \captionsetup{width=.9\linewidth}
    \captionsetup{justification=centering}
    \includegraphics[width=0.9\linewidth]{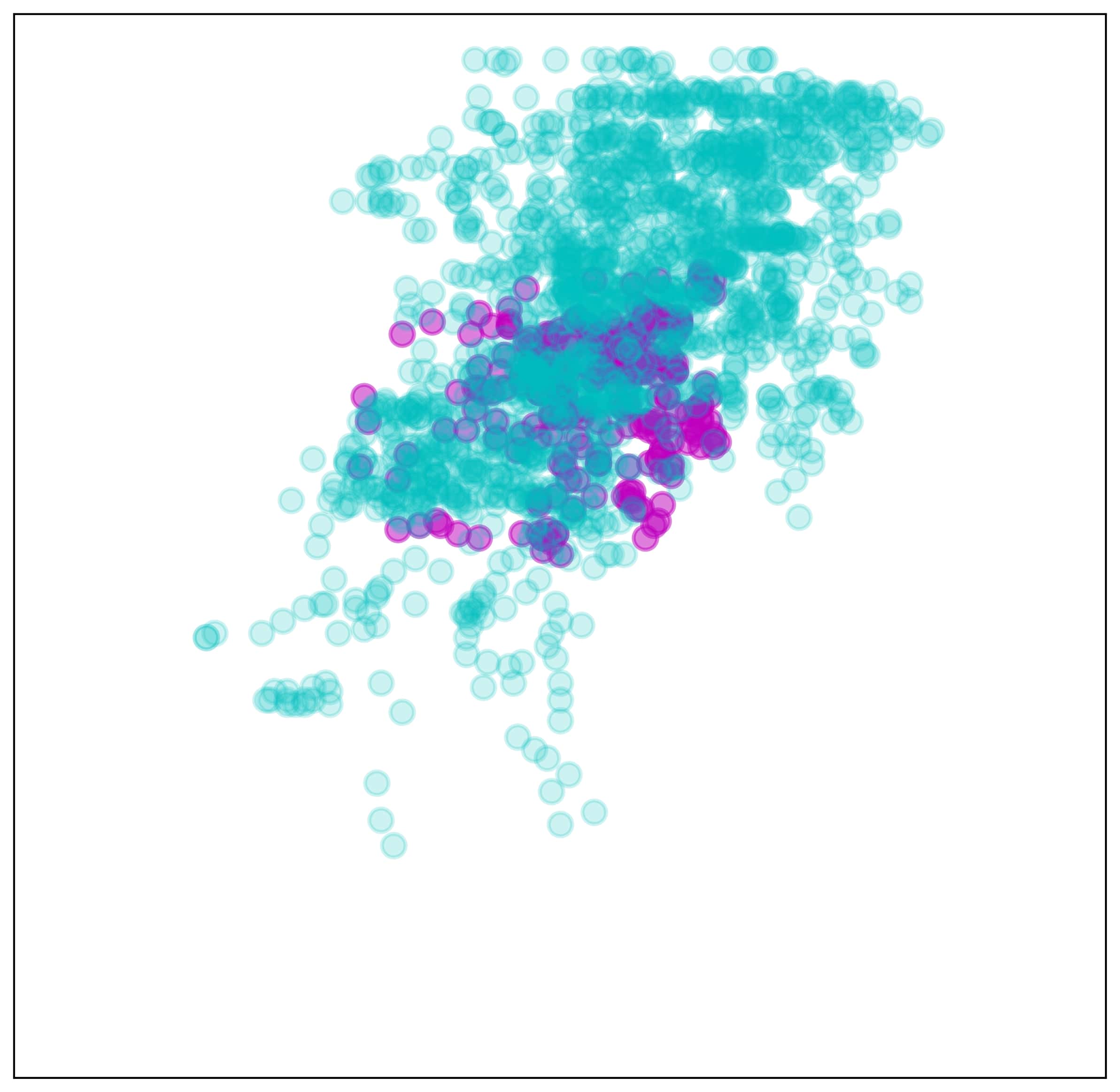}
    \caption{$19$.}
    \label{fig:ssm_19}
  \end{subfigure}%
  \begin{subfigure}[t]{0.25\linewidth}
      \centering
    \captionsetup{width=.9\linewidth}
    \captionsetup{justification=centering}
    \includegraphics[width=0.9\linewidth]{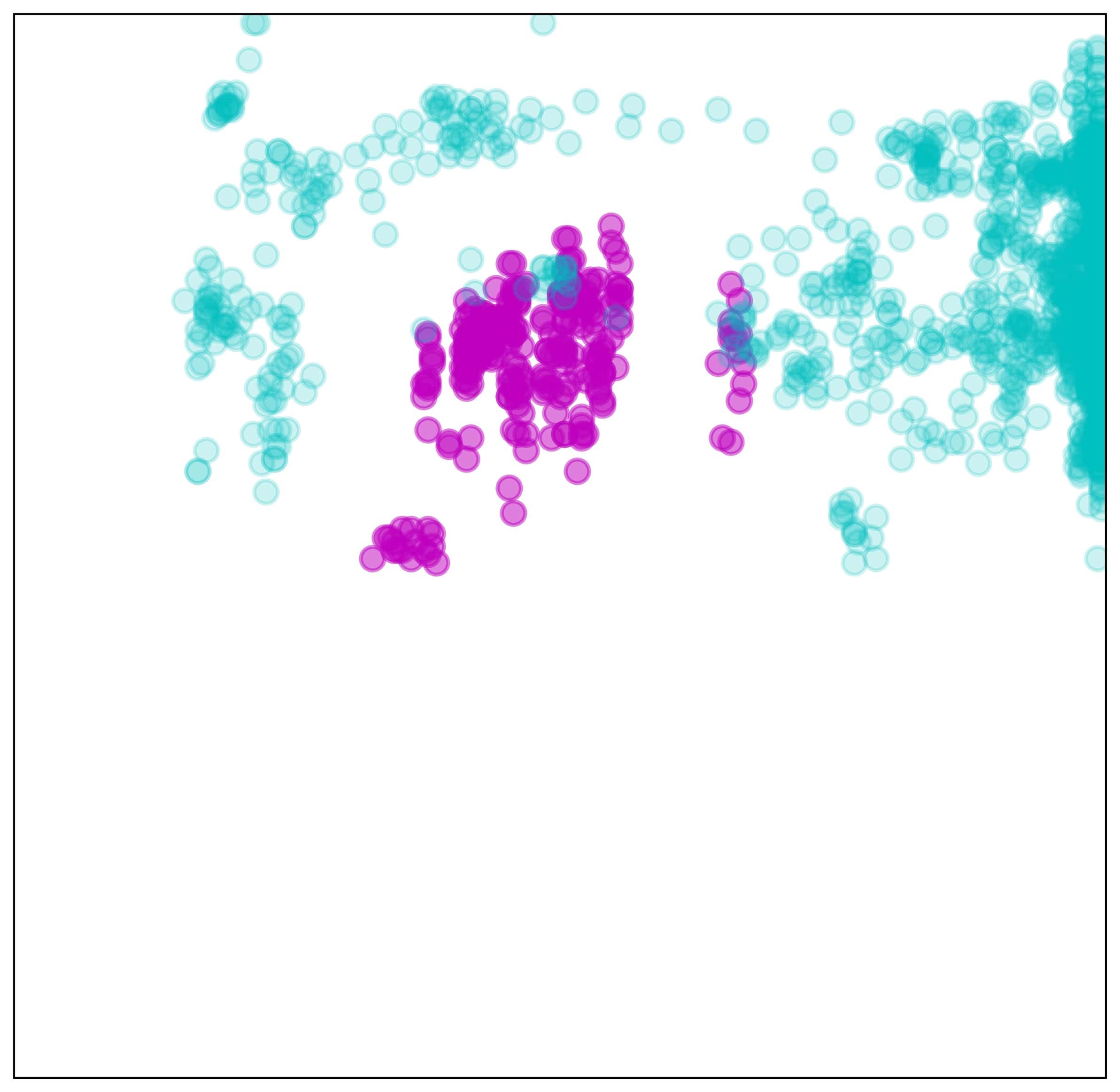}
    \caption{$2$.}
    \label{fig:ssm_2}
  \end{subfigure}%
  \begin{subfigure}[t]{0.25\linewidth}
      \centering
    \captionsetup{width=.9\linewidth}
    \captionsetup{justification=centering}
   \includegraphics[width=0.9\linewidth]{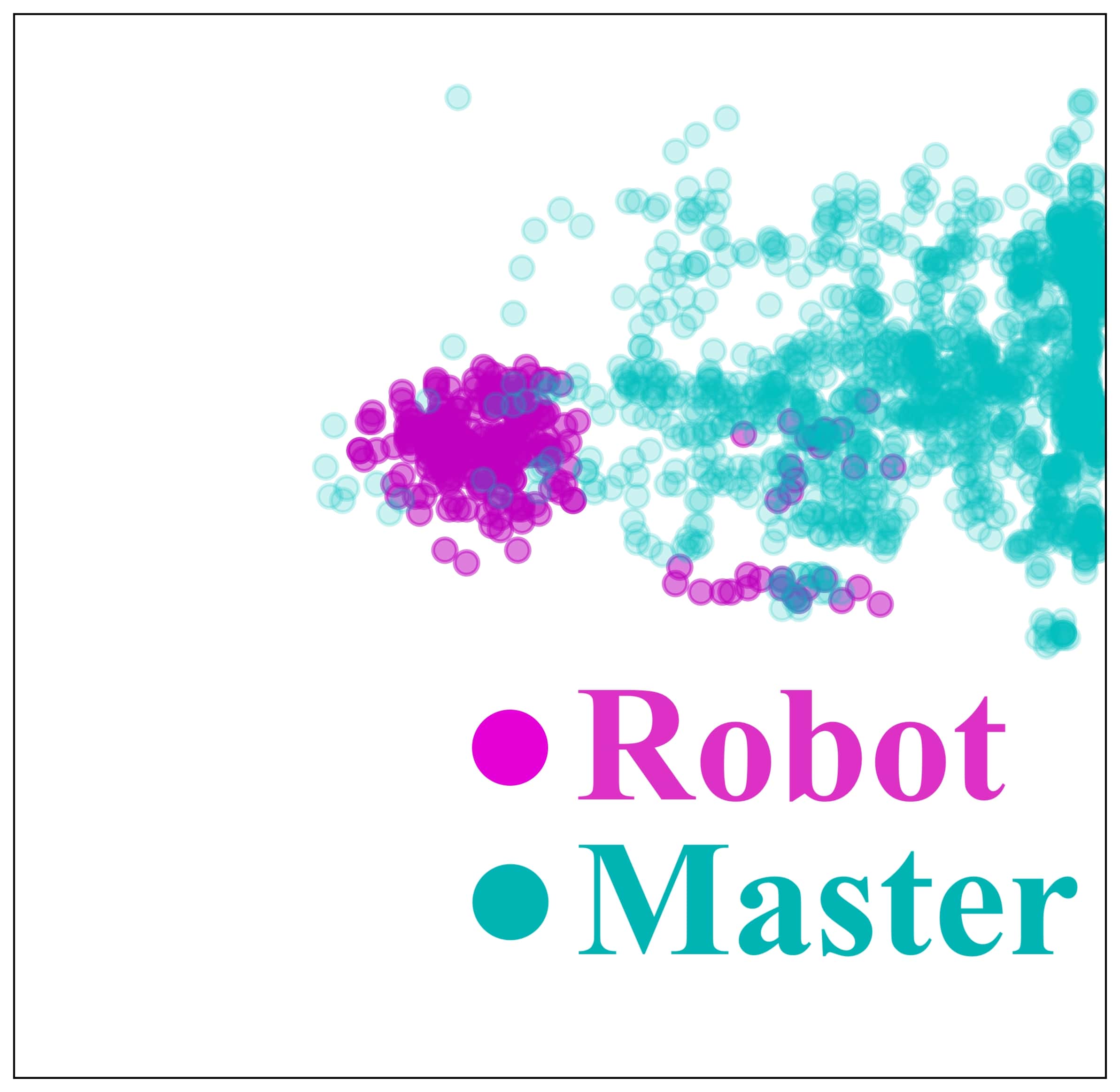}
   \caption{$30$.}
   \label{fig:ssm_30}
 \end{subfigure}%

  \caption{Visualized SpatialSoftmax feature locations on the master (validation set) and robot (trials executed by \textit{DA-force}). The number indicates the feature index (32 features in total). Some features overlapped between the robot and master ((\ref{fig:ssm_16}), (\ref{fig:ssm_19})), and other features predicted distinguished locations ((\ref{fig:ssm_2}), (\ref{fig:ssm_30})), which caused a disparity between policy predictions.}
  \label{fig:ssm_sample}
 \end{figure}
 
Fig. \ref{fig:success_example} illustrates that the proposed \textit{DA-force} successfully executed the bottle-cap-opening task on \textit{bottle-A}. Among all the tested model architectures, \textbf{\textit{DA-force}} had the best final cumulative success rate of $83.3\%$ by combining DA, F/T sensor input, and gaze (Fig. \ref{fig:success_rate}).

Second, \textbf{\textit{No-force}}, which did not use F/T sensory information, executed accurate reaching and grasping; however, poor manipulation during \textit{Rotate} caused a slip during cap opening (Fig. \ref{fig:failure_nosensor}), which resulted in a lower final success rate than \textit{DA-force}.

Third, \textbf{\textit{No-DA}} failed at accurately reaching the cap because it lacked the dual-action system (Fig. \ref{fig:failure_nodual}). Some may argue that the gripper touched the bottle cap; therefore, simply closing it may have led to successful grasping. However, the bottle collapsed when the gripper was closed manually.

Finally, \textbf{\textit{No-gaze}} demonstrated low accuracy for both \textit{GraspBottle} and \textit{GraspCap} (Fig. \ref{fig:failure_global}). This low accuracy was because the global vision-based policy was weak against distractions caused by the visual gap. 
The visualized example of two-dimensional visual features extracted by SpatialSoftmax on the global vision processing network demonstrated the difference in the feature coordinates between the demonstration data collected by the master (validation set) and the robot data collected by \textit{DA-force} during the test (Fig. \ref{fig:ssm_sample}). Some features focused on similar areas (Fig. \ref{fig:ssm_16}, \ref{fig:ssm_19}), whereas others focused on different areas (Fig. \ref{fig:ssm_2}, \ref{fig:ssm_30}). This difference resulted in errors in the action output computed from the MLP layer. Therefore the architecture using global vision failed to predict the accurate action.

\begin{figure}
  \centering
  \vspace{0.0in}
  \begin{subfigure}[t]{1.\linewidth}
    \centering
    \captionsetup{width=.9\linewidth}
    \captionsetup{justification=centering}
    \includegraphics[width=.9\linewidth]{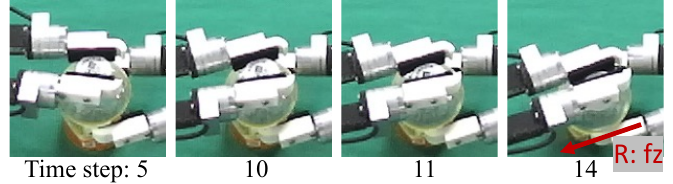}
    \caption{Visualized failure trial.}
    \label{fig:failure_release}
  \end{subfigure}%

  \vspace{0.0in}
  \begin{subfigure}[t]{0.5\linewidth}
    \captionsetup{width=.9\linewidth}
    \captionsetup{justification=centering}
    \includegraphics[width=.9\linewidth]{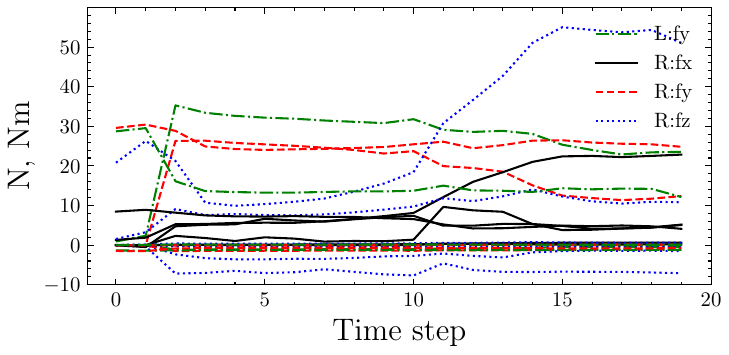}
    \caption{F/T sensor values for the failure trial.}
    \label{fig:sensor_fail}
  \end{subfigure}%
  \begin{subfigure}[t]{0.5\linewidth}
    \captionsetup{width=.9\linewidth}
    \captionsetup{justification=centering}
    \includegraphics[width=.9\linewidth]{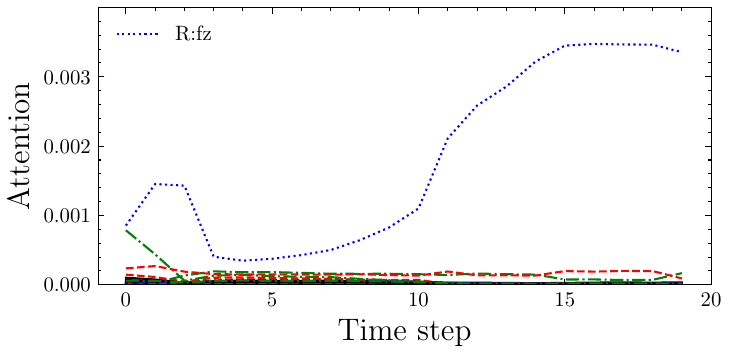}
    \caption{Attention for the failure trial.}
    \label{fig:att_fail}
  \end{subfigure}%
    \hfil
  \begin{subfigure}[t]{0.5\linewidth}
    \captionsetup{width=.9\linewidth}
    \captionsetup{justification=centering}
    \includegraphics[width=.9\linewidth]{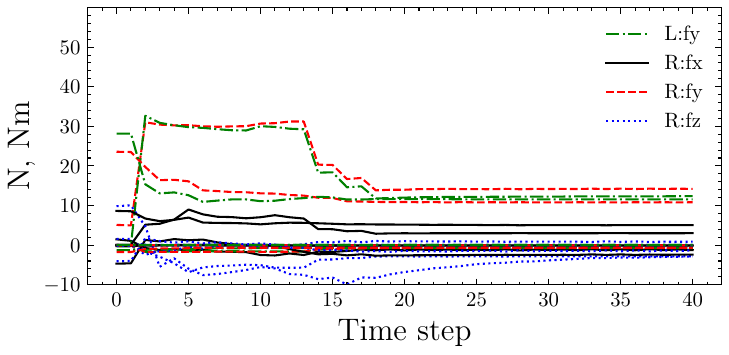}
    \caption{F/T sensor values for the successful trial.}
    \label{fig:sensor_succ}
  \end{subfigure}%
  \begin{subfigure}[t]{0.5\linewidth}
    \captionsetup{width=.9\linewidth}
    \captionsetup{justification=centering}
    \includegraphics[width=.9\linewidth]{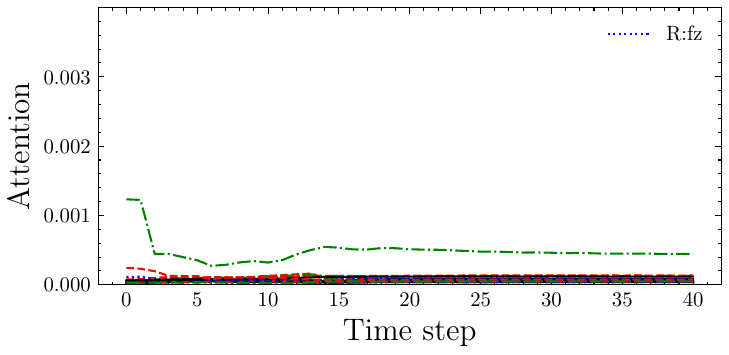}
    \caption{Attention for the successful trial.}
    \label{fig:att_succ}
  \end{subfigure}%

  \caption{Visualized F/T sensory values and attention of the Transformer for the sample successful and failure trials. R:fz refers to the force of the $z$-axis on the right arm.}
  \label{fig:att_res}
 \end{figure}

 Fig. \ref{fig:att_res} shows examples of F/T sensory values and the attention. First, sensory values of successful and failure trial were separately gathered from trials of \textit{DA-force} and \textit{No-force}. Then, the attention rollout \cite{abnar2020quantifying} of the Transformer from the force feedback network was computed. Then, the two-dimensional attention-rollout was averaged on each channel. The result demonstrates that if the robot failed to \textit{Rotate} (Fig. \ref{fig:failure_release}), the force of the $z$-axis on the right arm (R:fz) increased during rotation (Fig. \ref{fig:sensor_fail}, time step $5-11$), and the Transformer captured such an abnormality (Fig. \ref{fig:att_fail}). By contrast, in the successful trial, the force of R:fz did not increase (Fig. \ref{fig:sensor_succ}); therefore, the Transformer did not give any \textit{alert} (Fig. \ref{fig:att_succ}). Therefore, the F/T sensor provides prediction whether the robot would fail, and the proposed Transformer-based method can capture such alert.

 Table \ref{tab:success_rate_bottles} represents the results comparison of all bottles. Even though the \textit{bottle-B} and \textit{C} are trained with smaller number of dataset, the neural network well adapted to those new bottles.

\section{Discussion}
In this study, a master-to-robot policy transfer learning system using gaze-based dual-action deep imitation learning was demonstrated. This system is composed of hardware for the master-only demonstration, methods to reduce domain gaps, and gaze-based dual-action deep imitation learning with F/T sensory attention. 
The proposed system was tested on the bottle-cap-opening task for a real robot. In this task, the demonstration dataset was generated only from the master. The proposed method successfully learned bottle-cap opening with an $83.3\%$ success rate.

The proposed method demonstrated that the visual and kinematics gap can be suppressed using the gaze-based visual attention and the simple calibration.
The proposed system has a few advantages over other robot teaching methods. First, this system does not require a robot during the demonstration. Second, because demonstration data collection and training are possible from a simple controller that consists of encoders and F/T sensors, this demonstration system is comparatively low-cost compared with a bilateral teleoperation system. Third, a complex control scheme is not needed during the demonstration because the master controller is controlled directly by the human operator. Therefore, the proposed system is naturally safe, unlike a bilateral system that requires complex control schemes to achieve safety and stability. Finally, this M2R system can be scaled up efficiently. For example, a large-scale M2R system can consist of only one robot, and multiple masters can create large-scale demonstration data.

The method allows any gripper design provided it satisfies two conditions: (1) minimize human hand interference in the foveated vision, and (2) provide force feedback to the operator. For multi-DOF grippers, the design idea proposed in \cite{kanai2021third} may be beneficial.

One problem with the proposed system is that it requires cameras mounted in a fixed position on both the master and robot sides. This fixed camera mount restricts the robot's design to prevent intervention with the human's body. If the robot can learn from images acquired from different viewpoints, images from in front of the HMD will be available. This will be considered in future work.

\bibliographystyle{IEEEtran}
\bibliography{IEEEfull}

\begin{thebibliography}{10}
\providecommand{\url}[1]{#1}
\csname url@rmstyle\endcsname
\providecommand{\newblock}{\relax}
\providecommand{\bibinfo}[2]{#2}
\providecommand\BIBentrySTDinterwordspacing{\spaceskip=0pt\relax}
\providecommand\BIBentryALTinterwordstretchfactor{4}
\providecommand\BIBentryALTinterwordspacing{\spaceskip=\fontdimen2\font plus
\BIBentryALTinterwordstretchfactor\fontdimen3\font minus
  \fontdimen4\font\relax}
\providecommand\BIBforeignlanguage[2]{{%
\expandafter\ifx\csname l@#1\endcsname\relax
\typeout{** WARNING: IEEEtran.bst: No hyphenation pattern has been}%
\typeout{** loaded for the language `#1'. Using the pattern for}%
\typeout{** the default language instead.}%
\else
\language=\csname l@#1\endcsname
\fi
#2}}

\bibitem{kim2021gaze}
H.~{Kim}, Y.~{Ohmura}, and Y.~{Kuniyoshi}, ``Gaze-based dual resolution deep
  imitation learning for high-precision dexterous robot manipulation,''
  \emph{Robotics and Automation Letters}, pp. 1--1, 2021.

\bibitem{kim2021transformer}
H.~Kim, Y.~Ohmura, and Y.~Kuniyoshi, ``Transformer-based deep imitation
  learning for dual-arm robot manipulation,'' in \emph{International Conference
  on Intelligent Robots and Systems}, 2021.

\bibitem{zhang2018deep}
T.~Zhang, Z.~McCarthy, O.~Jow, D.~Lee, X.~Chen, K.~Goldberg, and P.~Abbeel,
  ``Deep imitation learning for complex manipulation tasks from virtual reality
  teleoperation,'' in \emph{International Conference on Robotics and
  Automation}, 2018, pp. 1--8.

\bibitem{kim2020using}
H.~Kim, Y.~Ohmura, and Y.~Kuniyoshi, ``Using human gaze to improve robustness
  against irrelevant objects in robot manipulation tasks,'' \emph{Robotics and
  Automation Letters}, vol.~5, no.~3, pp. 4415--4422, 2020.

\bibitem{hulin2011dlr}
T.~Hulin, K.~Hertkorn, P.~Kremer, S.~Sch{\"a}tzle, J.~Artigas, M.~Sagardia,
  F.~Zacharias, and C.~Preusche, ``The dlr bimanual haptic device with
  optimized workspace,'' in \emph{International Conference on Robotics and
  Automation}, 2011, pp. 3441--3442.

\bibitem{guo2019scaled}
J.~Guo, C.~Liu, and P.~Poignet, ``A scaled bilateral teleoperation system for
  robotic-assisted surgery with time delay,'' \emph{Journal of Intelligent \&
  Robotic Systems}, vol.~95, no.~1, pp. 165--192, 2019.

\bibitem{adachi2018imitation}
T.~Adachi, K.~Fujimoto, S.~Sakaino, and T.~Tsuji, ``Imitation learning for
  object manipulation based on position/force information using bilateral
  control,'' in \emph{2018 IEEE/RSJ International Conference on Intelligent
  Robots and Systems (IROS)}.\hskip 1em plus 0.5em minus 0.4em\relax IEEE,
  2018, pp. 3648--3653.

\bibitem{schou2013human}
C.~Schou, J.~S. Damgaard, S.~B{\o}gh, and O.~Madsen, ``Human-robot interface
  for instructing industrial tasks using kinesthetic teaching,'' in \emph{IEEE
  ISR 2013}.\hskip 1em plus 0.5em minus 0.4em\relax IEEE, 2013, pp. 1--6.

\bibitem{akgun2012trajectories}
B.~Akgun, M.~Cakmak, J.~W. Yoo, and A.~L. Thomaz, ``Trajectories and keyframes
  for kinesthetic teaching: A human-robot interaction perspective,'' in
  \emph{International Conference on Human-Robot Interaction}, 2012, pp.
  391--398.

\bibitem{Kuniyoshi1994learning}
Y.~Kuniyoshi, M.~Inaba, and H.~Inoue, ``{Learning by Watching: Extracting
  reusable task knowledge from visual observation of human performance},''
  \emph{Transactions on Robotics and Automation}, vol.~10, no.~6, pp. 799--822,
  1994.

\bibitem{yu2018one}
T.~Yu, C.~Finn, A.~Xie, S.~Dasari, T.~Zhang, P.~Abbeel, and S.~Levine,
  ``One-shot imitation from observing humans via domain-adaptive
  meta-learning,'' \emph{arXiv preprint arXiv:1802.01557}, 2018.

\bibitem{liu2018imitation}
Y.~Liu, A.~Gupta, P.~Abbeel, and S.~Levine, ``Imitation from observation:
  Learning to imitate behaviors from raw video via context translation,'' in
  \emph{International Conference on Robotics and Automation}, 2018, pp.
  1118--1125.

\bibitem{hayhoe2005eye}
M.~Hayhoe and D.~Ballard, ``Eye movements in natural behavior,'' \emph{Trends
  in Cognitive Sciences}, vol.~9, pp. 188--94, 2005.

\bibitem{vaswani2017attention}
A.~Vaswani, N.~Shazeer, N.~Parmar, J.~Uszkoreit, L.~Jones, A.~N. Gomez,
  {\L}.~Kaiser, and I.~Polosukhin, ``Attention is all you need,'' in
  \emph{Neural Information Processing Systems}, 2017, pp. 5998--6008.

\bibitem{pelz2001coordination}
J.~Pelz, M.~Hayhoe, and R.~Loeber, ``The coordination of eye, head, and hand
  movements in a natural task,'' \emph{Experimental Brain Research}, vol. 139,
  pp. 266--77, 2001.

\bibitem{bishop1994mixture}
C.~M. Bishop, ``Mixture density networks,'' Neural Computing Research Group,
  Aston University, Tech. Rep., 1994.

\bibitem{bazzani2016recurrent}
L.~Bazzani, H.~Larochelle, and L.~Torresani, ``Recurrent mixture density
  network for spatiotemporal visual attention,'' in \emph{International
  Conference on Learning Representations}, 2016.

\bibitem{finn2016deep}
C.~Finn, X.~Y. Tan, Y.~Duan, T.~Darrell, S.~Levine, and P.~Abbeel, ``Deep
  spatial autoencoders for visuomotor learning,'' in \emph{International
  Conference on Robotics and Automation}, 2016, pp. 512--519.

\bibitem{paillard1996fast}
J.~Paillard, ``Fast and slow feedback loops for the visual correction of
  spatial errors in a pointing task: a reappraisal,'' \emph{Canadian Journal of
  Physiology and Pharmacology}, vol.~74, no.~4, pp. 401--417, 1996.

\bibitem{kim2022robot}
H.~Kim, Y.~Ohmura, and Y.~Kuniyoshi, ``Robot peels banana with goal-conditioned
  dual-action deep imitation learning,'' \emph{arXiv preprint
  arXiv:2203.09749}, 2022.

\bibitem{he2016deep}
K.~He, X.~Zhang, S.~Ren, and J.~Sun, ``Deep residual learning for image
  recognition,'' in \emph{Conference on Computer Vision and Pattern
  Recognition}, 2016, pp. 770--778.

\bibitem{liu2019variance}
L.~Liu, H.~Jiang, P.~He, W.~Chen, X.~Liu, J.~Gao, and J.~Han, ``On the variance
  of the adaptive learning rate and beyond,'' in \emph{International Conference
  on Learning Representations}, 2020, pp. 1--14.

\bibitem{abnar2020quantifying}
S.~Abnar and W.~Zuidema, ``Quantifying attention flow in transformers,'' in
  \emph{Annual Meeting of the Association for Computational Linguistics}, 2020,
  pp. 4190--4197.

\bibitem{kanai2021third}
T.~Kanai, Y.~Ohmura, A.~Nagakubo, and Y.~Kuniyoshi, ``Third-party evaluation of
  robotic hand designs using a mechanical glove,'' \emph{arXiv preprint
  arXiv:2109.10501}, 2021.

\end{thebibliography}

\end{document}